\documentclass[11pt,a4paper]{article}
\usepackage[hyperref]{eacl2021}
\usepackage{times}
\usepackage{booktabs}
\usepackage{amsfonts}
\usepackage{multirow}
\usepackage{latexsym}
\usepackage{subcaption}
\usepackage{amsmath}
\usepackage{graphicx}

\DeclareMathOperator*{\argmax}{arg\,max}

\usepackage{microtype}

\aclfinalcopy 
\def\aclpaperid{662} 

\title{Multilingual Neural Machine Translation with \\Deep Encoder and Multiple Shallow Decoders}

\author{Xiang Kong$^{\ddagger}$\thanks{\ \ Work done at Facebook AI.} , Adithya Renduchintala$^{\dagger}$, James Cross$^{\dagger}$,\\{\bf Yuqing Tang$^{\dagger}$, Jiatao Gu$^{\dagger}$, Xian Li$^{\dagger}$ }\\
 $^{\ddagger}$Language Technologies Institute,
Carnegie Mellon University\\
$^{\dagger}$Facebook AI\\
{\tt xiangk@cs.cmu.edu}\\
    {\tt \{adirendu,jcross,yuqtang,jgu,xianl\}@fb.com}
}

\date{}

\begin{document}
\maketitle

\begin{abstract}
Recent work in multilingual translation advances translation quality surpassing bilingual baselines using deep transformer models with increased capacity. However, the extra latency and memory costs introduced by this approach may make it unacceptable for efficiency-constrained applications. It has recently been shown for bilingual translation that using a deep encoder and shallow decoder (DESD) can reduce inference latency while maintaining translation quality, so we study similar speed-accuracy trade-offs for multilingual translation. We find that for many-to-one translation we can indeed increase decoder speed without sacrificing quality using this approach, but for one-to-many translation, shallow decoders cause a clear quality drop. To ameliorate this drop, we propose a deep encoder with \emph{multiple} shallow decoders (DEMSD) where each shallow decoder is responsible for a disjoint subset of target languages. Specifically, the DEMSD model with 2-layer decoders is able to obtain a \textbf{1.8x speedup} on average compared to a standard transformer model with no drop in translation quality.
\end{abstract}
\section{Introduction}
Encoder-decoder based neural machine translation (NMT) systems have achieved great success on bilingual translation tasks ~\cite{sutskever2014sequence,cho2014learning,bahdanau2014neural,gehring2017convolutional,vaswani2017attention}. Recently, multilingual neural machine translation (MNMT) has also attracted much attention because of its ease of deployment, knowledge transfer among languages and the potential to enable zero-shot translation~\cite{dong-etal-2015-multi,firat2016multi,ha2016toward,johnson2017google,arivazhagan2019massively,zhang2020improving}. 
While MNMT can support translations in several directions, not all of them have better performance when compared to their corresponding bilingual models. Suspecting that poor performance in some directions is due to the limited model capacity, many prior works adopt deeper encoder and decoder ~\cite{zhang2019improving,wang2019learning,zhang2020improving}.
However, increasing the number of layers, especially in the decoder, deteriorates the latency of translation and memory costs. Recently, \newcite{kasai2020deep} show that given a fixed capacity budget, as measured by the number of layers, models with a deep encoder and a shallow decoder (DESD) are faster at inference time when compared to standard models with an equal number of encoder and decoder layers while maintaining translation quality.

Inspired by findings from \newcite{kasai2020deep}, in this work, we explore the speed-accuracy trade-off in multilingual machine translation systems. Given the same model capacity budget, we experiment various layer allocation strategies. We analyze multilingual models in the one-to-many (O2M) setting and  many-to-one (M2O) setting.  In the one-to-many setting, there are numerous target languages from a single source language (limited to English in this study); and in the many-to-one setting, several possible source languages are translated into a single target language (again, English in this study).

In the many-to-one scenario, we find that allocating more capacity to the encoder reduces the latency while achieving comparable performance.
We hypothesize that a deeper encoder helps the model accommodate multiple source languages,  while a shallow decoder is sufficient to support a single target language. 

However, in the one-to-many translation setting, speed-accuracy trade-off is complicated. We observe a performance drop as the decoder depth is reduced. We hypothesize that the shallow decoder can no longer model several different target languages adequately. With the goal of obtaining low latency while maintaining translation quality, we propose using \emph{multiple} shallow decoders where each decoder is responsible for a subset of the target languages.
Clearly, the introduction of multiple shallow decoders increases the size of our model. However, at inference time  only one shallow decoder will be used, thus not adding latency or memory costs. With multiple target languages and decoders, one natural question is how to assign each target language to one of these decoders.
We investigate several methods to assign each target language to one of these shallow decoders. 
More details are in the Section~\ref{sec:la}. Experimental results on three multilingual translation corpora show the effectiveness of our method to improve translation accuracy with lower latency at the same time.

Our main contributions are summarized as follows:
\begin{itemize}
    \item  We extend the speed-accuracy trade-off study of DESD models from bilingual to multilingual machine translation tasks with various layer allocations.
    \item We show that on many-to-one translation, multilingual DESD models enable 1.8x speedup on average without sacrificing performance comparing to the baseline (equal model capacity).
    \item We further proposed shared encoder and \emph{multiple} shallow decoders (DEMSD) for one-to-many setting again achieving 1.8x speed-up in decoding while preserving high-quality translations at the same time.

\end{itemize}

\begin{figure*}
     \centering
     \begin{subfigure}[b]{0.33\textwidth}
         \centering
         \includegraphics[width=\textwidth]{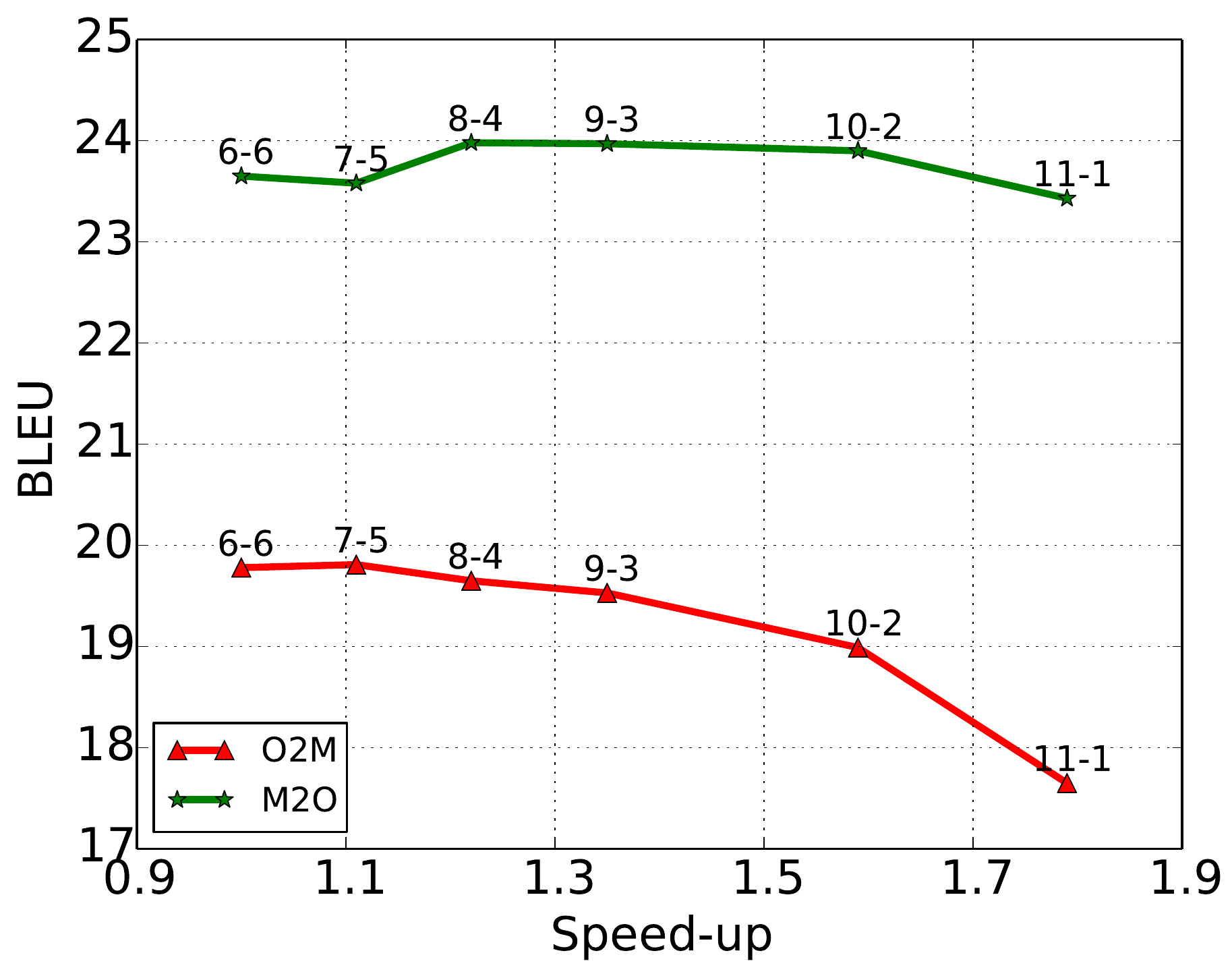}
         \caption{ML50}
     \end{subfigure}
     \begin{subfigure}[b]{0.32\textwidth}
         \centering
         \includegraphics[width=\textwidth]{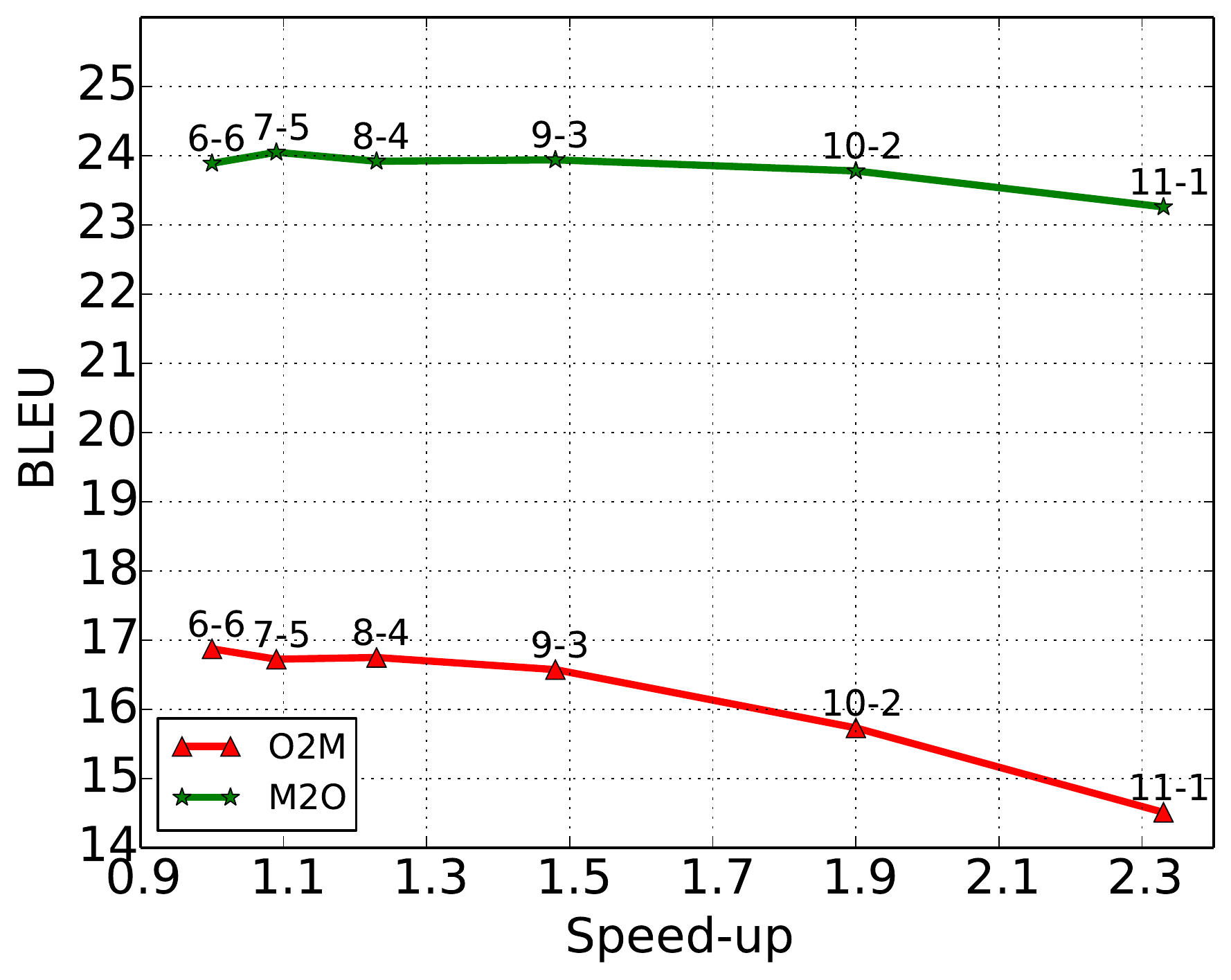}
         \caption{TED8-Related}
     \end{subfigure}
     \begin{subfigure}[b]{0.33\textwidth}
         \centering
         \includegraphics[width=\textwidth]{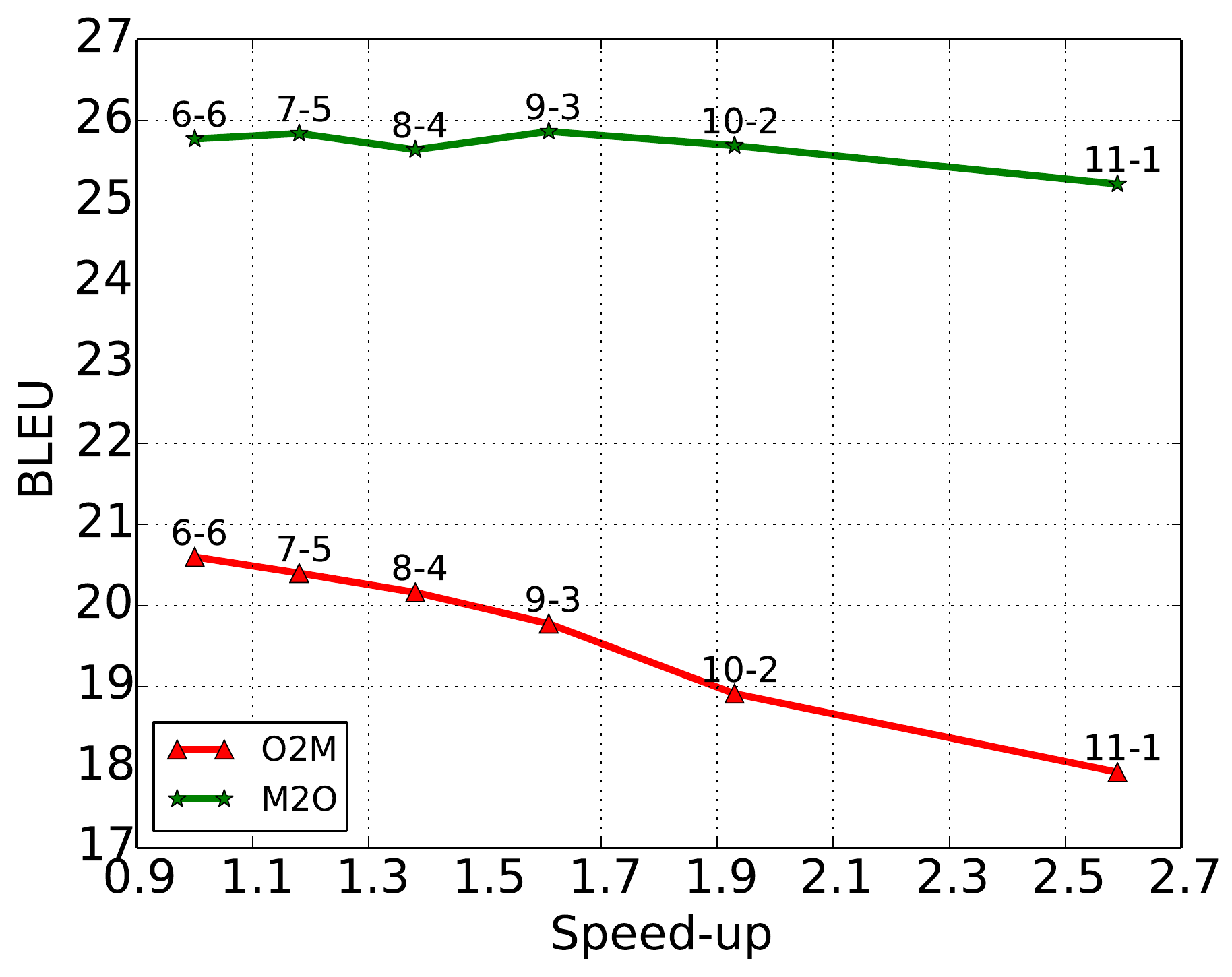}
         \caption{TED8-Diverse}
     \end{subfigure}
        \caption{Speed and accuracy trade-off of various layer allocations for O2M and M2O translations on ML50, TED8-Related and TED8-Diverse corpora. X-Y denotes X and Y layers in the encoder and decoder respectively. Best viewed in color.}
        \label{fig:trade-off}
\end{figure*}

\section{Deep encoder and shallow decoder (DESD) for multilingual NMT}

\paragraph{Background}
The transformer-based NMT model~\cite{vaswani2017attention} achieves state-of-the-art performance on many translation tasks. It consists of an encoder and a decoder, each of which contains several stacked layers. Since the transformer relies entirely on the attention mechanism, it allows more parallelization compared to recurrent neural networks. Specifically, at training time, the computation can be parallelized both in the encoder and decoder. At inference time, due to the auto-regressive property, the decoder needs to generate tokens one by one. However, the computation in the encoder is still parallelized given the source sentence. Therefore, the main latency of the transformer at inference time happens in the decoder, especially translating long sentences. Recently, \newcite{kasai2020deep} find that on bilingual machine translation tasks, putting more capacity of the transformer model to the encoder substantially reduces the decoding time and maintain the performance at the same time. 

Because this deep encoder and shallow decoder model achieves a superior speed-accuracy trade-off on bilingual translation tasks, in this section, we try to understand the layer allocations of transformer on the multilingual neural machine translation task given the same capacity budget which is measured by the number of layers in the encoder and decoder. We first experiment with three multilingual translation corpora.
\begin{itemize}
    \item ML50~\cite{tang2020multilingual}: a large-scale multilingual translation dataset which contains 49 languages$\leftrightarrow$English and more than 200 million training sentence pairs in total. All data are collected from open-resource data such as WMT, IWSLT, WAT, TED, etc.
    \item TED8-Related~\cite{wang2020balancing}: 4 low resource languages (Azerbaijani: az, Belarusian: be, Glacian: gl, Slovak: sk) and 4 relevant high resource languages (Turkish: tr, Russian: ru, Portuguese: pt, Czech: cs)
    \item TED8-Diverse~\cite{wang2020balancing}: 8 languages without consideration for relatedness (Bosnian: bs, Marathi: hr, Hindi: hi, Macedonian: mk, Greek: el, Bulgarian: bg, French: fr, Korean: ko)
\end{itemize}

Instead of just trying the shallowest possible decoder (1-layer), we train models with various configurations on each of these three corpora. Other than the layer allocation, all the other hyper-parameters and model configurations are the same among these models and the same training procedure is applied to these models (odel and training details are listed in Appendix~\ref{app:DESD_train}.). To understand the speed and accuracy trade-off of the layer allocation, two metrics are reported:
\begin{itemize}
    \item BLEU: the average tokenized BLEU score~\cite{papineni2002bleu} over all directions.
    \item DS: the decoding speed. It is measured by the number of tokens per second the system translates given one sentence at a time on a single GPU.
\end{itemize}

The results are shown in Figure~\ref{fig:trade-off}. Models with fewer decoder layers obtain higher decoding speed.
\subsection{Many-to-one translation}
In the M2O translation, there is no significant performance difference among these layer allocations. We hypothesize that this is because the deeper encoder learns better representations from a large number of source languages while on the decoder side only one language needs to be modeled. Therefore, given a more robust representation of source languages, the shallow decoder is able to generate high-quality translations. For example, the model with 10 encoder layers and 2 decoder layers obtain slightly better performance and a 1.8\textbf{x} speedup at the same time.

\subsection{One-to-many translation}
However, in the O2M translation setting, although models with the shallower decoder have lower latency compared to the standard transformer (6-6), there is a clear performance drop in terms of translation accuracy, especially for models with just 1 or 2 decoder layers. We attribute this to the shallow decoder not having enough capacity to model a large number of target languages. 
 
\section{Deep Encoder and Multiple Shallow Decoders (DEMSD)}
\label{sec:la}
We have seen that in one-to-many translation, DESD models have a performance drop compared to the standard transformer. In order to preserve translation quality and low latency at the same time, we propose a model with a shared encoder and \emph{multiple} shallow decoders (DEMSD), each of which is used to decode a subset of target languages. Although this will introduce more parameters, at inference time only one shallow decoder is needed for a given translation (since the output language is fixed) thus the model incurs no extra latency or memory costs.
One natural question that arises when using this multiple-decoder approach is how to assign output languages to each of the decoders. In this section, we explore several language assignment methods to assign each target language (or language group) to one of these multiple decoders. As a result, each decoder only needs to handle a disjoint subset of target languages.

\subsection{One language per decoder (EACH)}

The simplest way is to use a separate decoder for each output language. As a result, we will have as many decoders as the number of target languages and each decoder only needs to model one language.

\subsection{Random language set per decoder (RAND)}
In this method, we assign a random set of languages to a single decoder.
As the performance of the model will vary significantly based on the random assignment, we repeat this scheme with three different random assignments and report the average results. Instead of completely random grouping languages, we let each decoder handle a same number of languages but languages in one decoder are randomly grouped.

\subsection{One language family per decoder (FAM)}
Another intuitive way for language assignment is to use linguistic features~\cite{comrie1989language,lewis2009ethnologue,wals}, such as language family, typology, etc. In this method, we are guided by the intuition that languages from the same linguistic family share similar features which might be captured by a single decoder resulting in better performance. Thus, we group languages into several sets based on their linguistic families, and assign a family of languages to each decoder.
As a result, we will have as many decoders as the number of language families in the target languages. We expect that in the same decoder, a better knowledge transfer will happen among languages in the same language family. For example, in the TED8-Related corpus, 8 target languages are split into 4 languages families which are \textsc{Turkic}, \textsc{Slavic}, \textsc{Romance} and \textsc{Czech–Slovak}. The details are shown in Table~\ref{tab:lf}. The language family-based assignment results on other corpora are shown in Appendix~\ref{app:lf}.

\begin{table}[!t]
    \centering
    \begin{tabular}{c|c}
    \toprule
    Language Family & Languages\\
    \midrule
      \textsc{Turkic}   &  az, tr\\
        \textsc{Slavic}  &  be, ru\\
        \textsc{Romance} & gl, pt\\
        \textsc{Czech–Slovak} & sk, cs \\
    \bottomrule
    \end{tabular}
    \caption{Language families in the TED8-Related corpus.}
    \label{tab:lf}
    \vspace{-5mm}
\end{table}

\subsection{Pre-trained language embedding based assignment (EMB)}
From \newcite{johnson2017google}, a common way to indicate the target language is prepending a target language token to the source sentence. With the goal of capturing the information of languages they represent, their embeddings are trained end-to-end with source-target sentence pairs. We call these embeddings as the language embeddings here. According to \newcite{johnson2017google}, these language embeddings are able to capture target language features in their training data. Therefore, we first extract them from a well-trained model and group target languages according to them. Finally, each group are assigned to one of these decoders.

\subsection{Self-taught assignment (ST)} One disadvantage of the pre-trained language embedding based grouping method is the need of a pre-trained machine translation model. It would be better if the model assign each target languages to one of these multiple shallow decoders during the training automatically. We expect that given a fixed number of decoders and target languages, the model is capable of choosing the most appropriate decoders for each language. 

Specifically, our model consists of a shared encoder, $E$, and $N$ multiple decoders, $D=[D_1, D_2,...,D_N]$. Given a language $L$, the model will choose a decoder, $D_{i}$ for training and translation so that the log probability of output sequence $y$ given the input sequence $x$ is $\log p(y | x, E, D_{i})$ where $i$=$\argmax_{j}p(j|L_{e})$ and $p(\cdot|L_{e})$ is the probability of each decoder being chosen given the language $L$ and its language embedding vector $L_{e}$.
Intuitively, our model will learn the distribution of each decoder being chosen given a language and choose the one with the highest probability. However, the $\argmax$ operation here is non-differenetiable thus during trainiing we consider the Gumbel-Softmax~\cite{jang2016categorical}, a differentiable approximation of the $\argmax$ operation. 

In Gumbel-Softmax, it models the $p(j|L_{e})$ as:

\begin{equation}
    p(j|L_{e})=\frac{\exp(l_{j}+g_{j})/\tau}{\sum_{k=1}^{N}\exp(l_{k}+g_{k})/\tau}
\end{equation}
where $l$ is the logit and $g$=$-\log(-\log(u))$ and $u\sim \mathcal{U}(0, 1)$. In the forward pass, the differentiable approximation of the $\argmax$ operation is used to choose the decoder for the input language and during the backward,  the true gradient of the 
Straight-Through Gumbel-Softmax outputs is used.  In our experiments, the temperature $\tau$ is linearly reduced from 5 to 0.5. Finally, during training, the probability of the target sequence $y$ given the source sentence $x$ and multiple decoders $D$ is:
\begin{equation}
    p(y|x) = \sum_{n=1}^{N} p(n|L_{e})p_{n}(y|x)
\end{equation}
where $p_{n}(y|x)$ is the probability of $y$ given x in the n-th decoder and $p(n|L_{e})$ is the probability of the n-th decoder being sampled given the embedding of the language $L$. 
During inference, only the decoder with the highest probability will be used to decode the input sentences.

\section{Experiments}
\subsection{Experimental setup}
We conduct experiments on ML50, TED8-Related, TED8-Diverse multilingual machine translation corpora. ML50~\cite{tang2020multilingual} is an English-central translation benchmark of 50 languages with publicly available training and evaluation sets, including high, mid, and extremely low resource directions. Following~\newcite{tang2020multilingual}, we adopt the 250k SentencePiece model~\cite{kudo2018sentencepiece} used in XLM-R~\cite{conneau2019unsupervised} to tokenize the dataset so that all languages share the same vocabulary. For TED8-Related, TED8-Diverse corpora, we follow the preprocessing steps in~\newcite{wang2020balancing}.
\paragraph{Hyperparameters} 
On ML50, we follow most of the standard hyperparameters in the transformer-base~\cite{vaswani2017attention}: 8 attention heads per layer, 512 model dimensions, 2048 hidden dimensions and 0.1 dropout.  We train batches of 64k tokens using Adam~\cite{kingma2014adam} with
$\beta$ = (0.9, 0.98) and $\epsilon$ = $10^{-6}$ and 0.1 label smoothing. The learning rate goes to $1\mathrm{e}{-3}$ within 4,000 steps, and then decays with the inverse square-root schedule. All models are trained for 100,000 steps.
Furthermore, to mitigate the training data imbalance issue, the temperature sampling method is adopted~\cite{arivazhagan2019massively} which is set as 5 in all experiments.

On TED8 corpora, a smaller transformer model with 512 model dimensions, 1024 hidden dimensions and 0.3 dropout is adopted. All models are trained for 40k steps with batches of 16k tokens with a smaller learning rate $2\mathrm{e}{-4}$. The other training procedure is the same as the ML50.
\begin{table}[t]

\begin{subtable}[h]{0.5\textwidth}
        \centering
        \begin{tabular}{l|l}
        \toprule
        Method & Abbrev. \\
        \midrule
        One language per decoder &  EACH \\
        Random language set  &  RAND \\
        One language family per decoder & FAM \\
        Pre-trained Language embedding & EMB \\
        Self-taught & ST \\
        \bottomrule
        \end{tabular}
       \caption{The abbreviations of language assignment methods.}
       \label{tab:ann1}
    \end{subtable}
    \hfill
    \begin{subtable}[h]{0.5\textwidth}
        \centering
        \begin{tabular}{l|l}
        \toprule
        Metric & Abbrev. \\
        \midrule
        \# parameters at training time &  \#TP \\
        \# parameters at inference time & \#DP \\
        Decoding speed & DS \\
        \# decoders & \#DEC\\
    \bottomrule
        \end{tabular}
       \caption{The abbreviations of metrics.}
       \label{tab:ann2}
    \end{subtable}
    \caption{The abbreviations of language assignment methods and metrics.}
    \label{tab:abbrev_method}
    \vspace{-5mm}
\end{table}

\paragraph{Evaluation metrics}  For all models, we evaluate on the checkpoint with the best validation loss and use beam size 5 and length penalty 1.0 in decoding. Besides reporting the average BLEU score over all languages, on ML50, we predefine high ($>$ 1M), mid (100K, 1M]) and low ($<$ 100K) resource languages according to their data sizes and average BLEU scores on each of them are also computed. For the evaluation speed, \textbf{DS}, it is measured by the number of tokens the system translates per second given one sentence at a time on a single GPU.

\begin{table*}[t]
    \centering
    
    \begin{tabular}{clccccccccc}
    \toprule
        \# & Model  & BLEU & $\text{BLEU}_{\text{H}}$ & $\text{BLEU}_{\text{M}}$ & $\text{BLEU}_{\text{L}}$ & DS &  \#TP (M) & \#DP (M) & \#Dec\\
        \midrule
        1 & 6-6 & 19.68 & 19.60 & 18.99 & \bf 20.34 & 1.0\textbf{x} & 172 & 172 & 1\\
        \midrule
        2 &11-1 &  17.65 & 16.93 & 17.02 & 18.70 & 1.8\textbf{x} & 167 & 167 & 1\\
        3 &11-1-EACH & 17.83 & 19.18 & 18.32 & 16.47 & 1.8\textbf{x} & 368 &  167 & 49\\
        4 &11-1-RAND & 17.96 & 17.96 & 17.54 & 18.17 & 1.8\textbf{x} & 230 & 167 & 15\\
        5 &11-1-FAM & 18.34 & 18.25 & 17.92 & 18.79 & 1.8\textbf{x} & 230 & 167 & 15\\
        6 &11-1-EMB & 18.19 & 18.17 & 17.79 & 18.40 & 1.8\textbf{x} & 230 & 167 & 15\\
        7 & 11-1-ST & 18.47 &18.31 & 18.02 & 18.79 & 1.8\textbf{x} & 230 & 167 & 15\\
        \midrule
        8 &10-2 &  18.99 & 18.50 & 18.41 & 19.72 & 1.6\textbf{x} & 168 & 168 & 1\\
        9 &10-2-EACH & 18.93 & \bf 21.45 & \bf 19.42 & 16.78 & 1.6\textbf{x} & 572 & 168 & 49 \\
        10 &10-2-RAND & 19.24 & 19.64 & 18.76 & 19.24 & 1.6\textbf{x} & 294 & 168 & 15\\
        11 &10-2-FAM & 19.70 & 20.01 & 19.31 & 19.81 & 1.6\textbf{x} & 294 & 168 & 15\\
        12 &10-2-EMB & 19.64 & 20.08 & 19.01 & 19.94 & 1.6\textbf{x} & 294 & 168 & 15\\
        13 &10-2-ST & \bf 19.71 & 20.03 & 19.36 & 19.98 & 1.6\textbf{x} & 294 & 168 & 15\\
        \bottomrule
    \end{tabular}
    \caption{Comparison among various models on ML50. $\text{BLEU}_{\text{H}}$, $\text{BLEU}_{\text{M}}$ and $\text{BLEU}_{\text{L}}$ denote the average BLEU score over high, mid and low resource languages respectively. More notation information can be found in Table~\ref{tab:abbrev_method}}
    \label{tab:main_res_ml50}

\end{table*}

\begin{table*}[t]
    \centering
    \small
    \begin{tabular}{cl|ccccc|ccccc}
    \toprule
        \multirow{2}{*}{\#} & \multirow{2}{*}{Model} & \multicolumn{5}{c}{Related} & \multicolumn{5}{c}{Diverse} \\
        && \#TP (M) & \#DP (M)  & \#Dec & DS & BLEU &  \#TP (M) & \#DP (M)  & \#Dec & DS & BLEU  \\
        \midrule
        1 & 6-6 & 64 & 64 & 1 & 1.0\textbf{x} & 16.75 & 66 & 66 & 1 & 1.0\textbf{x} & 20.60 \\
        \midrule
        2 & 11-1 & 58 & 58  & 1 & 2.3\textbf{x} & 14.51 & 61 & 61 & 1 & 2.6\textbf{x}  & 17.94\\
        3 & 11-1-EACH & 80 & 58 & 8 & 2.3\textbf{x} & 14.81 & 83 & 61 & 8 & 2.6\textbf{x} & 18.68 \\
        4 & 11-1-RAND & 58 & 58 & 4 & 2.3\textbf{x} & 14.69 & 74 & 61 & 5 & 2.6\textbf{x} & 18.37 \\
        5 & 11-1-FAM & 68 & 58  &  4 & 2.3\textbf{x} & 15.20 & 74 & 61 & 5 & 2.6\textbf{x}  & \bf 18.82\\
        6 & 11-1-EMB & 68 & 58  &  4 & 2.3\textbf{x} & 15.20 & 74 & 61 & 5 & 2.6\textbf{x}  & 18.62\\
        7 & 11-1-ST & 65 & 58 & 3 & 2.3\textbf{x} & 15.04 & 74 & 61 & 5 & 2.6\textbf{x}  & 18.64\\
        \midrule
        8 & 10-2 & 59 & 59 & 1 & 1.9\textbf{x} & 15.73 & 62 & 62 & 1 & 1.9\textbf{x} & 18.91 \\
        9 & 10-2-EACH & 104 & 59 & 8 & 1.9\textbf{x} &  16.13 & 106 & 62 & 8 & 1.9\textbf{x} & 19.60 \\
        10 & 10-2-RAND & 78 & 59  & 4 & 1.9\textbf{x} & 16.04 & 87 & 62 & 5 & 1.9\textbf{x} & 19.23\\
        11 & 10-2-FAM & 78 & 59  & 4 & 1.9\textbf{x} & \bf 16.88 & 87 & 62 & 5 & 1.9\textbf{x} & \bf 20.25 \\
        12 & 10-2-EMB &78 & 59  & 4 & 1.9\textbf{x} & \bf 16.88 & 87 & 62 & 5 & 1.9\textbf{x} & 19.98\\
        13 & 10-2-ST & 78 & 59  & 4 & 1.9\textbf{x} & 16.71 &
        87 & 62 & 5 & 1.9\textbf{x} & 20.03\\
    \bottomrule
    \end{tabular}
    \caption{Translation speed and accuracy trade-off on TED8-Related and TED8-Diverse corpora. Notation information can be found in Table~\ref{tab:abbrev_method}.}
    \vspace{-5mm}
    \label{tab:main_res_ted}
\end{table*}
\begin{figure}[t]
    \centering
    \includegraphics[width=0.5\textwidth]{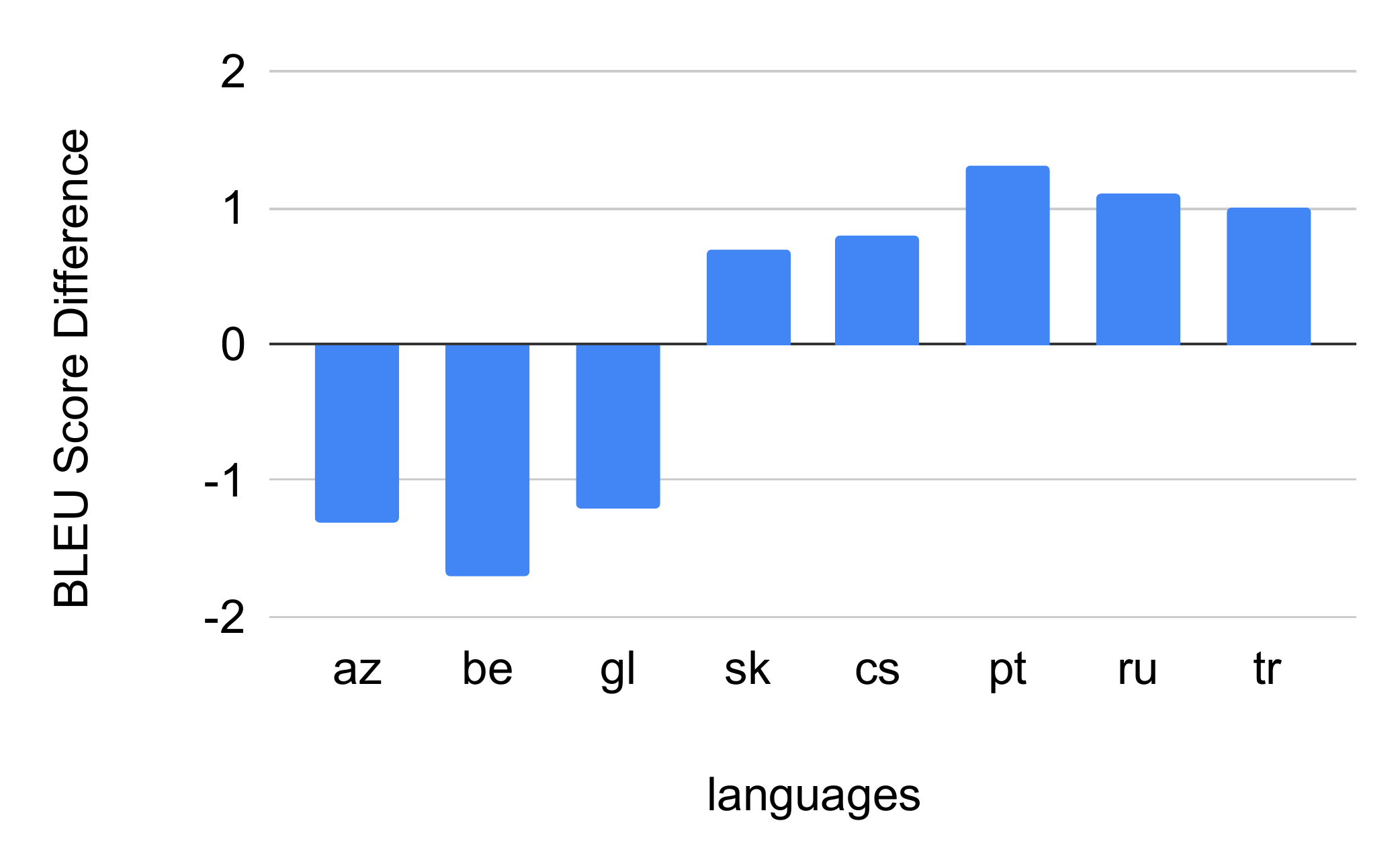}
    \caption{The BLEU score difference between models 10-2-EACH and 10-2 on TED8-Related ($\textrm{BLEU}_\textrm{{10-2-EACH}}-\textrm{BLEU}_\textrm{{10-2}}$). (Left four languages are low-resourced and the right four are high-resourced.)}
    \label{fig:sl_diff}
    \vspace{-5mm}
\end{figure}
\subsection{Results}

From Figure~\ref{fig:trade-off}, we find that for O2M translation, models with 1- or 2-layer decoders have a clear performance drop compared to the standard transformer (6-6). Therefore, our main experiments adopt multiple shallow decoders with 1 and 2 decoder layers. Results on ML50 and TED8 corpora are shown in Table~\ref{tab:main_res_ml50} and~\ref{tab:main_res_ted} respectively. For simplicity, we introduce the abbreviation of each language assignment method and evaluating metrics in Table~\ref{tab:abbrev_method}.

\paragraph{One language per decoder (EACH)} With this assignment method, models obtain superior performance on high and mid resource languages but poor results on low resource languages.
On ML50, if each language has its own decoder, we find that it achieves great results on high resource languages ($\text{BLEU}_{\text{H}}$ in rows 2 vs. 3 and 8 vs. 9 in Table~\ref{tab:main_res_ml50}). We think that given enough training data, the shallow decoder has enough ability to model one language. However, it performs worse on the low resource languages compared with the baseline ($\text{BLEU}_{\text{L}}$ between rows 2 vs. 3 and 8 vs. 9 in Table~\ref{tab:main_res_ml50}). To further understand this assignment method, we also show the BLEU score differences between models 10-2 and 10-2-EACH on TED8-Related in Figure~\ref{fig:sl_diff}. The left three languages are relatively low resourced and their performance is lower than the baseline model in which all languages share one decoder\footnote{Note that although sk is defined as a low resourced language in this dataset, the reason why language sk still have slightly better result is that sk has 61.5k training data but the other three low resource languages (az, be, gl) have less than 10k training sentence pairs.}.  This also demonstrates that their decoders are not able to learn robust representations given a limited amount of training data. And decoders trained with high resource languages generate higher quality translations and we attribute this to the enough training data and no negative transfer effect when trained without other languages~\cite{arivazhagan2019massively}.

\begin{figure}[t]
    \centering
    \includegraphics[width=0.5\textwidth]{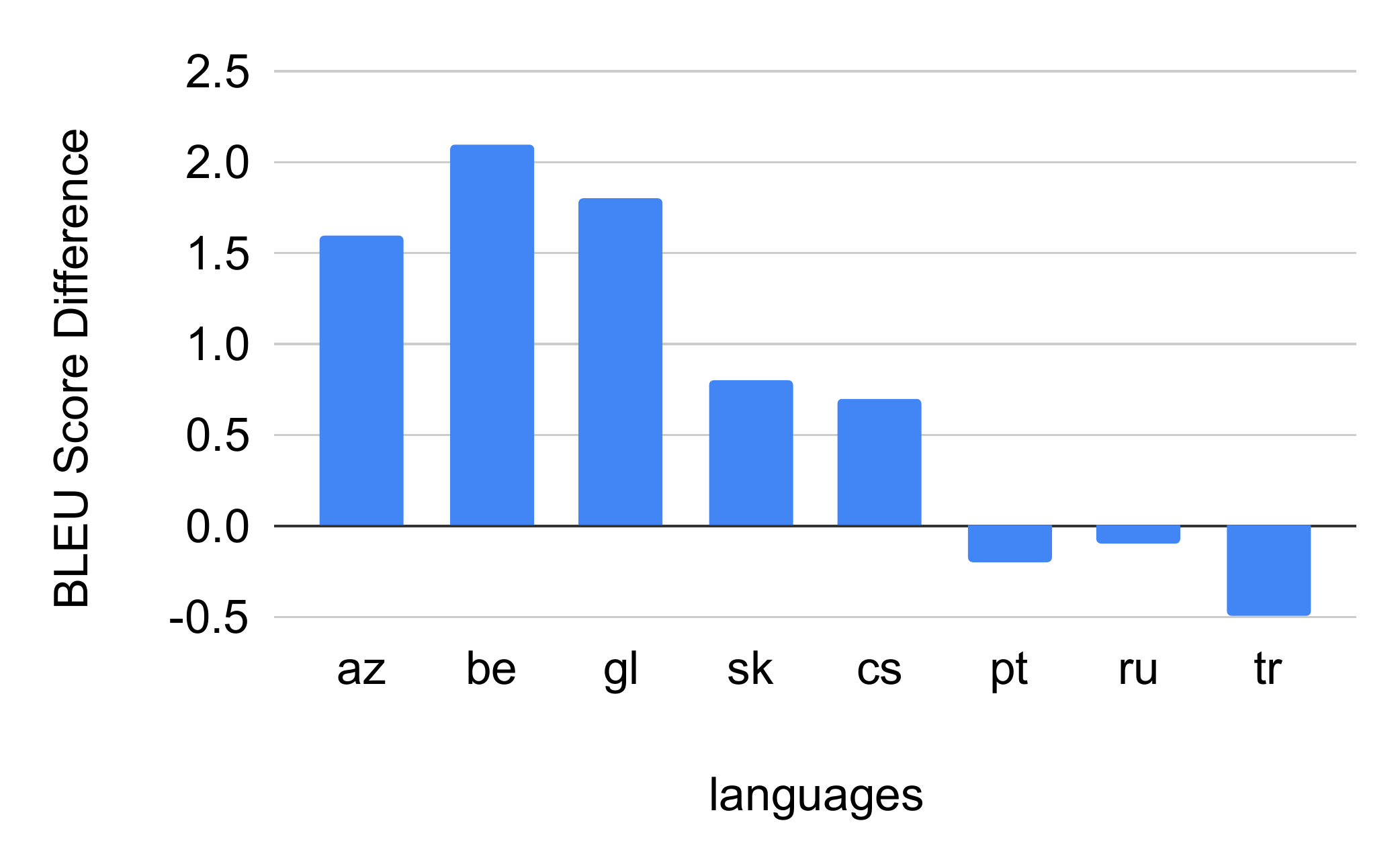}
    \caption{The BLEU score difference between models 10-2-FAM and 10-2-EACH on TED8-Related ($\textrm{BLEU}_\textrm{{10-2-FAM}}-\textrm{BLEU}_\textrm{{10-2-EACH}}$). (Left four languages are low-resourced and the right four are high-resourced.)}
    \label{fig:llf_diff}
\end{figure}

\begin{figure*}[!ht]
     \centering
     \begin{subfigure}[b]{0.48\textwidth}
         \centering
         \includegraphics[width=\textwidth]{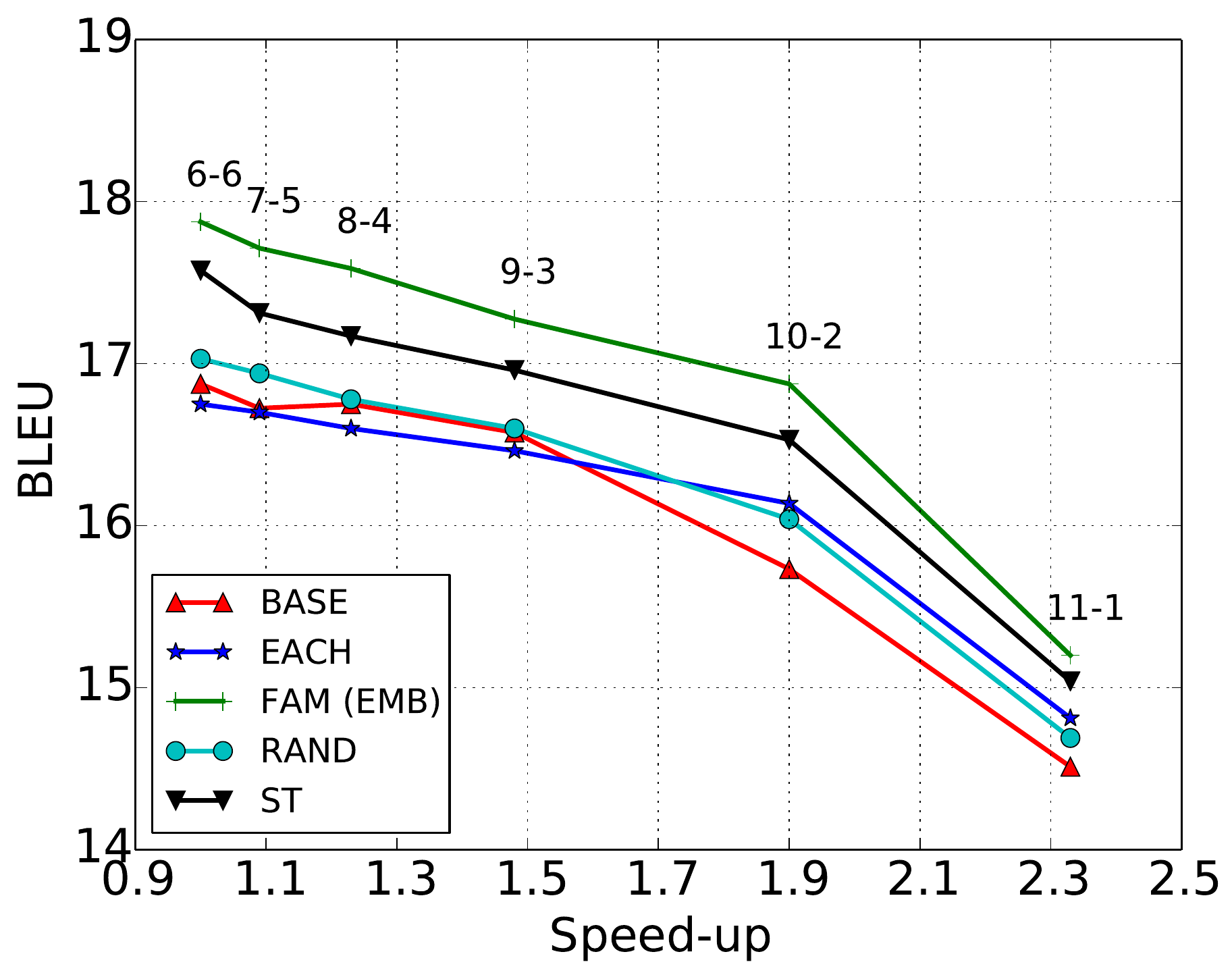}
         \caption{TED8-Related}
     \end{subfigure}
     \hfill
     \begin{subfigure}[b]{0.48\textwidth}
         \centering
         \includegraphics[width=\textwidth]{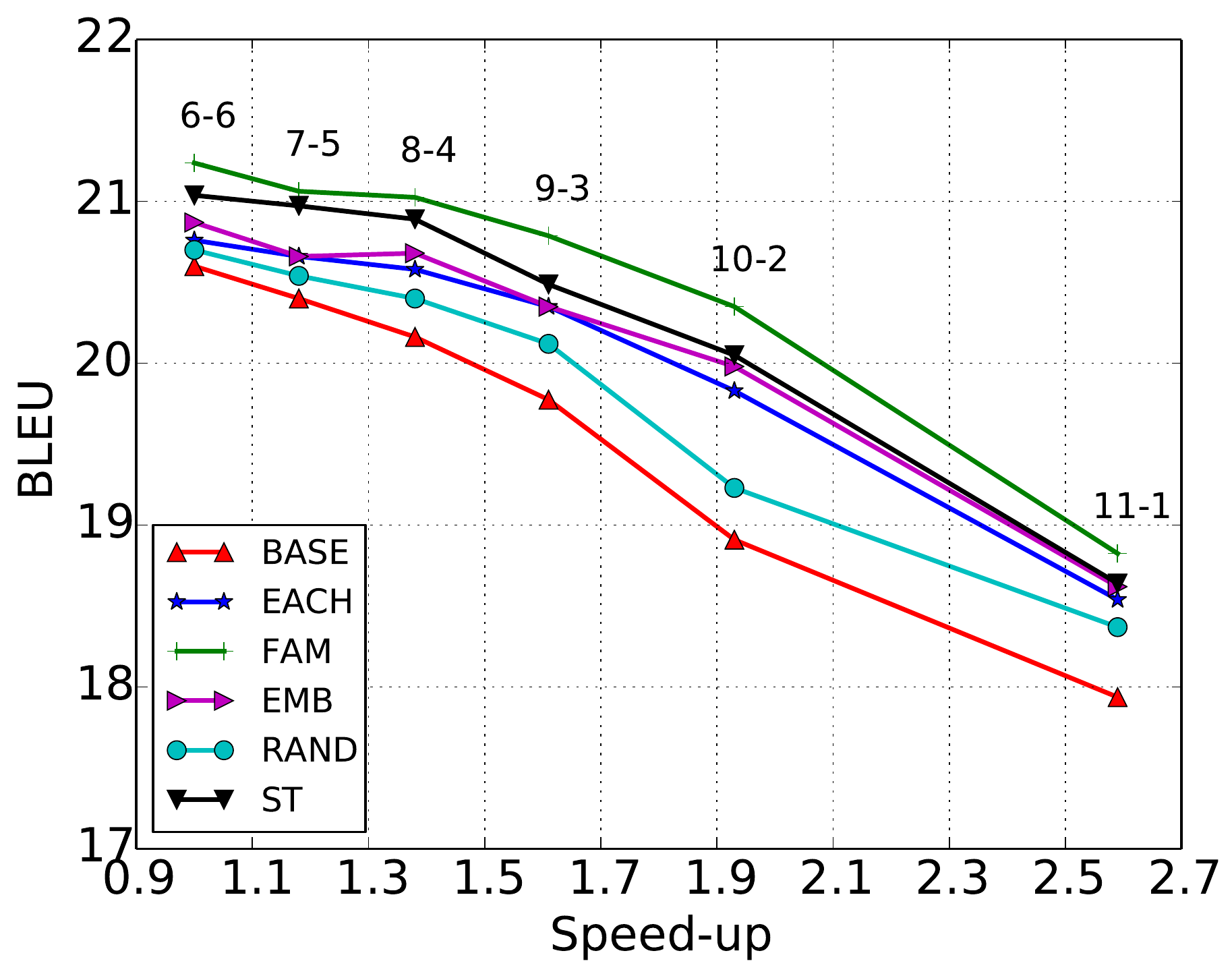}
         \caption{ TED8-Diverse}
     \end{subfigure}
        \caption{Multiple decoders with various layer allocations of Transformer on ML50, TED8-Related and TED8-Diverse corpora. X-Y denotes X and Y layers in the encoder and decoder respectively. 'BASE' denotes the shared decoder model.}
        \label{fig:trade-off-full}
        \vspace{-5mm}
\end{figure*}

\paragraph{Random language set assignment (RAND)} We find that random language set assignment slightly improve the performance over the baseline due to the sub-optimal knowledge transfer among languages in the same decoder.
If each decoder handles a similar number of languages, it also slightly improve the performance compared to the model with one shared decoder (BLEU scores between rows 2 vs. 4 and 8 vs. 10 in Tables~\ref{tab:main_res_ml50} and \ref{tab:main_res_ted}). We attribute this to that the shallow decoder performs better given fewer languages. This also demonstrates that one shallow decoder does not have enough capacity to model a large number of languages. However, compared to language family and embedding assignment methods, the random language set method has lower translation quality, showing that how to assign target languages into these decoders is also crucial.

\paragraph{One language family per decoder (FAM)}
We group all languages into several groups according to their language families and assign each family to one shallow decoder. As a result, we have 15, 4, 5 language families in ML50, TED8-Related and TED8-Diverse corpora respectively. From the comparison between rows 2 vs. 5 and 8 vs 11 in Tables~\ref{tab:main_res_ml50} and \ref{tab:main_res_ted}. It is clear to find that language family-based decoders achieves better accuracy and maintain the low latency at the same time. Furthermore, for models with multiple 2-layer decoders, they achieve comparable performance with the model 6-6 and obtain around a 1.8 times speedup at inference time. We think the improvement is mainly coming from the better knowledge transfer among similar languages (in one language family).
In order to understand this further, we plot the BLEU score difference between models 10-2-EACH and 10-2-FAM on TED8-Related in Figure~\ref{fig:llf_diff}.

We find that the major improvement of model 10-2-FAM over 10-2-EACH is from the low resource languages which means the high resource languages help their relevant low resource languages effectively. 

\paragraph{Language embedding-based assignment (EMB)}
For the fair comparison, languages are also grouped into the same number of language families according to language embeddings from the well-trained baseline model 6-6. Grouping results are listed in the Appendix~\ref{app:le}. We first find that language embedding-based grouping method is able to group similar languages together, showing the ability of language embeddings to effectively capture language characteristics during training. For example, on TED8-Related, the language embedding achieve the same grouping result as the language family-based one shown in Table~\ref{tab:lf}. The language embedding-based assignment method achieves similar results compared to the language family-based one and effectively improve the performance of the baseline model.

\paragraph{Self-taught language assignment (ST)}
In this method, the model tries to assign target languages to multiple decoders automatically and there is no need having any prior knowledge (linguistic families) or well-trained models (language embeddings). From the rows 7 vs. 2 and 13 vs. 8 in Tables~\ref{tab:main_res_ml50} and~\ref{tab:main_res_ted}, our self-taught method improves around 1 BLEU score over the baseline. It also achieves similar results compared with the language family (embedding)-based language assignment methods, demonstrating the effectiveness of this method.

\section{Analysis and Discussion}
\subsection{Multiple decoders for various layer allocations}
In our main experiments, we use multiple very shallow decoders (i.e., 1 and 2-layer decoders) because there is a clear performance drop when using a single decoder with this configuration for one-to-many translation compared with the standard transformer (6-6), and compared to deeper decoders, employing multiple 1- or 2-layer decoders keeps the number of parameters manageable at training time. Nevertheless, it will be meaningful to explore the effect of multiple decoders on various layer allocations. Considering the model size and tractable training time, we only conduct experiments on TED8 corpora and the results are shown in Figure~\ref{fig:trade-off-full}. On each line (the same language assignment method), the deeper decoders achieve better performance and the shallower decoder has lower latency. Moreover, if we compare language family-based assignment and the baseline models, given the same decoding speed at inference time, the former one consistently improve the performance with the same decoding speed at inference time. And with the similar performance, e.g., 10-2-FAM and 6-6, our best multiple shallow decoder models have much lower latency. 

\subsection{Speed-accuracy trade-off in multilingual machine translation}
From the above experiments and findings, in the one-to-many translation, the DESD framework obtains superior speed-accuracy trade-off.  For example, the model with 10 encoder layers and 2 decoder layers obtain slightly better accuracy and a 1.8\textbf{x} speedup.

Under the one-to-many setting, multiple shallow decoders are needed to mitigate the performance drop of the DESD model. And the crucial part is to group languages with similar features to one decoder to obtain the better knowledge transfer among languages (our FAM, EMB and ST methods). With this, our DEMSD model with multiple 2-layer decoder is capable of achieving similar performance and a 1.8\textbf{x} speedup compared to the standard transformer.

\section{Related Work}
\label{sec:rw}
Speed and accuracy are two important metrics to evaluate a machine translation system. In this work, we mainly discuss the transformer architecture~\cite{vaswani2017attention}. A number of works have explored various ways to improve its inference speed. \newcite{kim-etal-2019-research} adopt shallow decoder and layer trying to speed up the inference on CPUs. \newcite{shi-knight-2017-speeding} and \newcite{senellart-etal-2018-opennmt} employee vocabulary reduction to speed up the softmax layer. \newcite{li2020deep} employ a latent depth transformer model which prune layers during inference time to reduce the inference cost. There are also some works optimizing attention computations to speed up the inference speed~\cite{zhang-etal-2018-accelerating,kitaev2020reformer,katharopoulos_et_al_2020,chelba2020faster}. Recently, \newcite{kasai2020deep} places more capacity to the encoder side and keep an extremely shallow (one-layer) decoder to achieve a superior speed-accuracy trade-off.

Multilingual neural machine translation (MNMT) is an attractive field recently~\cite{firat2016multi,ha2016toward,johnson2017google} because MNMT tries to employ one model to translate more than one language pair, even including ones unseen during training (zero-shot translation). Knowledge transfer among languages boosts the performance of low-resource languages. However, many works~\cite{arivazhagan2019massively,zhang2020improving,aharoni2019massively} have shown the capacity bottleneck of translation when modeling many languages. Therefore, before simply stacking more layers in the encoder and decoder, it is crucial to first understand how to balance the speed and accuracy given a fixed capacity budget. Therefore, in this work, we try to understand various capacity allocations to achieve the best speed-accuracy trade-off.

\section{Conclusion}
In this work, we study speed-accuracy trade-offs using various layer configurations for multilingual neural machine translation. We find that for many-to-one translation, deep encoder and shallow decoder (DESD) models improve decoding speed while maintaining translation quality with the same model capacity.
However, for one-to-many translation we do observe a drop in quality when the decoder depth is reduced.
To mitigate the performance drop of DESD models in one-to-many translation, we proposed using a shared encoder and \emph{multiple} shallow decoders (DEMSD). Our best DEMSD models with 2-layer decoders are capable of speeding up decoding by 1.8 times while achieving the same quality compared to a standard transformer. 

Our work can be combined with techniques mentioned in Section~\ref{sec:rw} such as optimized attention computation, vocabulary reduction, knowledge distillation, etc. We expect that these combinations will further improve the decoding speed and obtain a better speed-accuracy trade-off. This work can also be extended to other encoder-decoder applications beyond translation, such as question answering, dialogues, and so on. We will explore these directions in the future work.

\bibliography{anthology,eacl2021}
\bibliographystyle{acl_natbib}

\clearpage
\appendix

\section{Training details of DESD model}
\label{app:DESD_train}
In order to explore how DESD models work on multilingual machine translation, we train transformer-based models with various layer allocations on three multilingual machine translation corpora, ML50, TED8-Related and TED8-Diverse. For the fair comparison, the training process is the same across all models.

On ML50, we employ the standard transformer-base model: 8 attention heads per layer, 512 model dimensions, 2048 hidden dimensions and 0.1 dropout. All models are trained for 100,000 with batches of 64k tokens using Adam and 0.1 label smoothing. The learning rate goes to $1\mathrm{e}{-3}$ within 4,000 steps,and then decays with the inverse square-root schedule. 

On TED8 corpora, following~\cite{wang2020balancing}, a smaller transformer model is adopted, i.e., 4 attention heads per layer, 512 model dimensions, 1024 hidden dimensions and 0.3 dropout. All models are trained for 40,000 with batches of 16k tokens using Adam and 0.1 label smoothing. The learning rate goes to $2\mathrm{e}{-4}$  within 4,000 steps, and then decays with the inverse square-root schedule. 

\section{Language family assignment results}
\label{app:lf}
In Table~\ref{tab:lf_td}, we show the language family-based assignment result on TED8-Diverse. Since this corpus is collected without considering relatedness, some groups just have one language. But its multiple decoders model improves the accuracy, showing the effectiveness of this method.

The language families in ML50 is shown in Table~\ref{tab:lf_ml50}.
\begin{table}[!t]
    \centering
    \begin{tabular}{c|c}
    \toprule
    Language Family & Languages\\
    \midrule
      \textsc{Indo-iranian}   &  hi, hr\\
        \textsc{Slavic}  &  mk, bs, bg\\
        \textsc{Korean} & ko \\
        \textsc{Hellenic} & el \\
        \textsc{Romance} & fr \\
    \bottomrule
    \end{tabular}
    \caption{Language families in the TED8-Related corpus.}
    \label{tab:lf_td}
\end{table}

\begin{table}[!ht]
    \centering
    \begin{tabular}{c|c}
    \toprule
    Language Family & Languages\\
    \midrule
      \textsc{Dravidian}   &  ta, ml, te, vi\\
        \textsc{Germanic}  &  de, nl, sv, af\\
        \textsc{Indo-Aryan} & hi, si, ne, ar, ur, mr, gu, bn \\
        \textsc{Iranian} & ps, fa \\
        \textsc{Chinese} & zh \\
        \textsc{Baltic} & lv \\
        \textsc{Austroasiatic} & km, id, xh, he \\ 
        \textsc{Japonic} & ja \\
        \textsc{Koreanic} & ko \\
        \textsc{Kra-Dai} & th, my \\
        \textsc{Pollsh} & pl \\
        \textsc{Romance} & fr, es, lt, ro, it, pt, gl \\
        \textsc{Slavic} & ka, mn, cs, ru, hr, uk, mk, sl \\
        \textsc{Turkic} & tr, kk, az \\
        \textsc{Uralic} & fi, et \\
    \bottomrule
    \end{tabular}
    \caption{Language families in the TED8-Related corpus.}
    \label{tab:lf_ml50}
\end{table}

\section{Language embedding assignment results}
\label{app:le}
On TED8-Related, we obtain the same language assignment results as the language family-based one.
On TED8-Diverse, the result of language embedding assignment is pretty similar to the language family assignment result (Table~\ref{tab:le_td}. The only difference is that language bg is grouped with language mk. We think this is because the language embedding not only contains the linguistic feature but the data feature as well.

\begin{table}[!ht]
    \centering
    \begin{tabular}{c|c}
    \toprule
    Group Id & Languages\\
    \midrule
     0 & hi, ko\\
     1 & mk, bg \\
     2 & ko, bs \\
     3 & el \\
     4 & fr \\
    \bottomrule
    \end{tabular}
    \caption{Language embedding-based language assignment result on the ML50 corpus.}
    \label{tab:le_td}
\end{table}

\begin{table}[t]
    \centering
    \begin{tabular}{c|c}
    \toprule
    Group Id & Languages\\
    \midrule
      0   &  ta, ml, te\\
       1 &  de, nl, af\\
        2 & hi, si, ne, ur, mr, gu, bn \\
        3 & ps, fa \\
        4 & zh \\
       5& lv \\
       6 & km, id, xh, he \\ 
        7 & ja \\
        8& ko \\
       9 & th, my, vi \\
       10& pl, sv, ar  \\
        11 & fr, es, lt, ro, it, pt, gl \\
        12& ka, mn, cs, ru, hr, uk, mk, sl \\
        13& tr, kk, az \\
        14 & fi, et \\
    \bottomrule
    \end{tabular}
    \caption{Language embedding-based grouping results on ML50.}
    \label{tab:le_ml50}
\end{table}
\end{document}